\documentclass{article}

\usepackage[dandb]{neurips_2025}
\usepackage[dvipsnames]{xcolor}         
\usepackage{graphicx}
\usepackage{multirow}
\usepackage{subcaption}
\usepackage{tcolorbox}

\usepackage[utf8]{inputenc} 
\usepackage[T1]{fontenc}    
\usepackage{hyperref}       
\usepackage{url}            
\usepackage{booktabs}       
\usepackage{amsfonts}       
\usepackage{amsmath}       
\usepackage{nicefrac}       
\usepackage{microtype}      

\title{A Manually Annotated Image-Caption Dataset for Detecting Children in the Wild}

\author{%
  Klim Kireev\thanks{Equal contribution.} \\
  MPI-SP \& EPFL \\
  Bochum, Germany \\
  \texttt{klim.kireev@mpi-sp.org} \\
  \And
  Ana-Maria Cre\c{t}u$^*$ \\
  EPFL\\
  Lausanne, Switzerland \\
  \texttt{ana-maria.cretu@epfl.ch} \\  
  \AND
  Raphael Meier \\
  Cyber-Defence Campus \\
  armasuisse S+T \\
  Thun, Switzerland \\
  \texttt{raphael.meier@armasuisse.ch} \\
  \And
  Sarah Adel Bargal \\
  Georgetown University \\
  Washington, D.C., USA \\
  \texttt{sarah.bargal@georgetown.edu} \\
  \And
  Elissa Redmiles \\
  Georgetown University \\
  Washington, D.C., USA \\
  \texttt{elissa.redmiles@georgetown.edu} \\
  \And 
  Carmela Troncoso \\
  MPI-SP \& EPFL \\
  Bochum, Germany \\
  \texttt{carmela.troncoso@mpi-sp.org} \\
}

\begin{document}

\maketitle

\begin{abstract}
Platforms and the law regulate digital content depicting minors (defined as individuals under 18 years of age) differently from other types of content. Given the sheer amount of content that needs to be assessed, machine learning-based automation tools are commonly used to detect content depicting minors. To our knowledge, no dataset or benchmark currently exists for detecting these identification methods in a multi-modal environment. To fill this gap, we release the Image-Caption Children in the Wild Dataset (ICCWD), an image-caption dataset aimed at benchmarking tools that detect depictions of minors.
Our dataset is richer than previous child image datasets, containing images of children in a variety of contexts, including fictional depictions and partially visible bodies. ICCWD contains 10,000 image-caption pairs manually labeled to indicate the presence or absence of a child in the image. To demonstrate the possible utility of our dataset, we use it to benchmark three different detectors, including a commercial age estimation system applied to images. Our results suggest that child detection is a challenging task, with the best method achieving a 75.3\% true positive rate. We hope the release of our dataset will aid in the design of better minor detection methods in a wide range of scenarios.
\end{abstract}

\setcounter{footnote}{0} 

\section{Introduction}
\label{sec:intro}
Most digital platforms restrict the sharing of content that relates to minors, defined as persons under 18 years of age\footnote{We use ``minor'' and ``child'' interchangeably, for any person under 18 years of age.}. Platforms may prohibit content showing abuse involving minors (Youtube~\cite{youtube2025}, WeChat~\cite{wechat2025}) or illegal behaviors like minors drinking or smoking (Youtube~\cite{youtube2025}, TikTok~\cite{tiktok2025}), or may impose restrictions on monitization of content depicting minors (TikTok~\cite{tiktok2025}, Instagram~\cite{abc2024}).

An even more restricted type of content depicting minors is Child Sexual Abuse Material (CSAM).

CSAM creation, possession, and sharing is illegal in most jurisdictions around the world~\cite{icmec2016}. In spite of this, the spread of CSAM online has been growing exponentially~\cite{bursztein2019rethinking}. 

The threat has further grown with bad actors using text-to-image models to generate AIG-CSAM directly or as a building block in downstream applications such as ``nudifying'' services~\cite{iwf2023}. Models such as StableDiffusion have been used to produce AI-generated CSAM (AIG-CSAM) on an unprecedented scale~\cite{rombach2021highresolution}. 
Whether or not AIG-CSAM relates to real children, it is illegal in several jurisdictions, such as the UK, and studies have shown that its spread over the Internet has negative effects on society~\cite{iwf2023}.
Current implemented and proposed strategies for model developers to prevent AIG-CSAM generation include prohibiting users from generating content that sexualizes children~\cite{midjourney2025,gemini2025} (and filtering the outputs of models to detect violations) and filtering children from training datasets of T2I models~\cite{thorn2024}.

It is common to resort to machine learning (ML) approaches to automatically detect whether a piece of content depicts a minor (throughout the paper, we will refer to such approaches as ``minor detection methods'')~\cite{sightengine2025,apilayer2025,sae2014towards,macedo2018benchmark,torres2019gender,islam2019forensic,chaves2020improving,rondeau2022deep,gangwar2021attm,moodley2023detecting}. 
Minor detection methods have many other applications~\cite{weda2007automatic,agarwal2018image,chua2019development,fitwi2020minor,laranjeira2022seeing}, such as anonymizing children's faces on CCTV footage~\cite{fitwi2020minor}, detecting the presence of children in cars with the aim of preventing vehicular heatstroke~\cite{chua2019development}, and tailoring advertising for children~\cite{weda2007automatic}.

To the best of our knowledge, the majority of minor detection methods in the image domain are implemented using face-based age estimation methods~\cite{phsmoura2020, bursztein2019rethinking}.
The typical workflow of a face-based age estimator is to use a face detector to detect faces, then infer the age for every detected face using ML.
Perhaps as a result, publicly-available datasets used to evaluate minor detection methods~\cite{chaves2020improving,phsmoura2020,fitwi2020minor,rondeau2022deep,moodley2023detecting,gangwar2021attm} only include images with faces, where the label (age or boolean child/not child) is assigned using the facial information.
They do not include images with labels assigned based on body parts where the face of the child is not visible, nor do they include images of children that are not photographs, such as graphic art, cartoons, and statues. However, restricting data to face-containing photographs does not align with regulations enforced by law or commercial platforms. These regulations stay in place if the face of the child is not visible, or even if the child is fictional in the case of CSAM. Further, novel methods that claim to use body information for age estimation are in development to improve gaps in the efficacy of face-based age estimators~\cite{kuprashevich2023mivolo}. Datasets that include body information are necessary to develop and evaluate such methods. 

Existing datasets also lack textual descriptions (captions), which means that they are not suitable for evaluating minor detection methods that use additional contextual information available to platforms or law enforcement, such as image descriptions or message texts. 

Nor would un-captioned datasets be optimal in evaluating minor detection methods designed with the purpose of filtering the training datasets of T2I models.
These models are trained on multi-modal, \textit{web-scale} and \textit{largely uncurated} datasets such as LAION-5B~\cite{schuhmann2022laion}.
Filtering children from these datasets has been proposed as a potential strategy to prevent harmful downstream uses such as AI-CSAM generation~\cite{thorn2024} and to prevent models from reproducing the likeness of real children, for privacy reasons~\cite{hrw2024australia}. To our knowledge, no dataset or benchmark is available to enable or evaluate the impact of such filtering.

\textbf{Contributions.} 
In this paper, we release Image-Caption Children in the Wild Dataset (ICCWD), a dataset designed for benchmarking minor detection methods in a multi-modal environment.
It is the first image-caption dataset for this problem, which means that, unlike the current datasets, it can be used to evaluate detection methods that utilize caption information. 
Moreover, ICCWD is also better suited for the evaluation of image-only detection methods, since it contains not only photos with distinctive faces, but also partially visible bodies, body parts, and other depictions of people (including minors) such as statues, graphic art, and cartoons. This scenario is especially important if the detector is used to enforce existing regulations, which generally hold regardless of the face visibility or origins of the image.
 
We make our dataset publicly available on HuggingFace\footnote{\url{https://huggingface.co/datasets/amcretu/iccwd}}.

As a proof-of-concept of the utility of our dataset, we use it to benchmark three minor detection methods: a caption-based classifier relying on DeepSeek-V3~\cite{liu2024deepseek}, a state-of-the-art LLM, and an image-based classifier relying on Amazon Rekognition Image~\cite{amazonrekognition}, a commercial face-based age estimation system, and a classifier combining the two.
We make our code available at GitHub\footnote{\url{https://github.com/spring-epfl/iccwd/}}.

Our results suggest that detecting content depicting children, when not restricted to facial images of children, is a challenging task, with the best method achieving 75.3\% true positive rate on our proposed challenging and comprehensive multi-modal dataset. 
We hope that our study initiates further development of robust methods to detect content depicting children.

\section{Related Work}
\label{sec:related-works}

To our knowledge, there are no publicly available image-caption datasets for evaluating minor detection methods. Here, we overview the two types of image-only datasets suitable for evaluating minor detection methods. 

\textbf{Image-only minor detection datasets.}
We are aware of five publicly available image-only datasets suitable for detecting content depicting minors, none of which have captions available: \textit{Juvenile-80k}~\cite{gangwar2021attm} (80k images), \textit{Child Image Detection}~\cite{phsmoura2020} (4.8k images), \textit{YLFW} ~\cite{ylfw2024} (10k images), \textit{HDA-SynChildFaces}~\cite{falkenberg2024syn} (188k), and \textit{Children's Face Dataset}~\cite{styleganchilddataset} (10k images). All three datasets contain only images with faces, and therefore are not suitable for benchmarking detection methods when the face is not available. In addition to that, \textit{Juvenile-80k}, \textit{YLFW}, and \textit{Child Image Detection} contain only photographs, while \textit{Children's Face Dataset} and \textit{HDA-SynChildFaces} contain only AI-generated images. Therefore, none of them have fictional (drawings, cartoons, etc.), but not AI-generated images. Our proposed dataset addresses both gaps.

\textbf{Age estimation datasets.} 
Other datasets used to evaluate minor detection methods are obtained by combining two or more image-only age estimation datasets~\cite{chaves2020improving,gangwar2021attm} such as IMDB-Wiki~\cite{rothe2015dex}, FG-Net~\cite{fu2014interestingness}, and UTKFace~\cite{zhang2017age}. Age estimation datasets~\cite{kuprashevich2023mivolo,gallagher2009understanding,fu2014interestingness,eidinger2014age,rothe2015dex,zhang2017age,moschoglou2017agedb,agustsson2017apparent,karkkainen2021fairface} use face-based labeling, meaning that they do not include images with age labels for people whose face is not visible in the image. Furthermore, some age estimation data sets are labeled with age buckets that include the age of 18, such as 10-19~\cite{gallagher2009understanding,karkkainen2021fairface} and 15-20~\cite{eidinger2014age}, making it impossible to use their labels to distinguish minors from adults.

Finally, these datasets contain only photographs with one notable exception of Szasz et al~\cite{szasz2022measuring}, who created a manually curated dataset of children's book illustrations labeled with coarse age labels (infant, child, teenager, adult, and senior).

Even though this dataset contains only facial images and is relatively small (980 images), the authors show that age estimation methods trained on datasets of human faces are suboptimal when applied to illustrations, which supports the necessity of datasets that contain both real and fictional images for better benchmarking, as our proposed dataset does.

\section{Dataset: ICCWD}
\label{sec:dataset}
We propose Image-Caption Children in the Wild Dataset (ICCWD), a dataset of image-caption pairs manually labeled with whether the image contains a depiction of a child, resulting in 1,675 child images. 
The intended use of this dataset is to evaluate minor detection methods -- including methods that take into account both image and text information -- when applied to photo-realistic and/or generated content, fictional content, and content that depicts children's bodies but not their faces . 

\subsection{Dataset Curation} 

The dataset consists of 10,000 entries (1,675 labeled as Child) with the following attributes: \texttt{URL, caption, label, num\_people, sha256\_hash, pdq\_hash}.

The \texttt{URL} and \texttt{caption} pairs are sourced from Google's Conceptual Captions-3M (CC3M) dataset~\cite{sharma2018conceptual}.
CC3M is an image-caption dataset of 3.3M image samples with high-quality captions.  
It was built by processing a large number of webpages to extract images together with Alt-text HTML attributes as the captions.
The resulting images were extensively filtered for quality and image-caption alignment.
Furthermore, captions were generalized through a set of transformations, e.g., dates were removed and named entities were replaced with hypernyms (e.g., ``Harisson Ford'' was replaced with ``actor''). Relying on well-curated captions guarantees that measuring performance using this dataset provides an upper bound of caption-based or caption-assisted minor detection, as detection methods will work worse on noisy captions. 

In September 2024, we downloaded all the training images of CC3M that were available at the provided
URLs using the  \texttt{img2dataset}\footnote{\url{https://github.com/rom1504/img2dataset}} library. 
This resulted in 2,267,817 samples (roughly 68\% of the original dataset).

\subsection{Dataset Annotation} 
To assess whether it was viable to randomly sample images for labeling directly from our CC3M sample, we compute the (rough) estimate of child images by counting captions with child-related keywords (``child'', ``baby'', ``kid'', ``infant'', ``toddler'', ``boy'', ``girl'' and their plurals). We obtain 120,410 samples (5.3\% of the downloaded samples).
This low prevalence of child images in the CC3M dataset implies that direct sampling from our CC3M sample would require us to label an excessively large number of images.

Thus, to reduce the number of images that we need to label, we applied a state-of-the-art object detector, YOLO-11~\cite{yolo11_ultralytics} to the downloaded images and filtered out all images that do not contain people. 
We chose YOLO-11 because this model is capable of detecting body parts and also fictional human depictions, such as cartoons and statues. This enables us to build a diverse dataset of human depictions not limited to photorealistic facial images. 

This filtering process leaves us with 951,217 samples, i.e., 42\% of the downloaded dataset.
We randomly sample 10,000 of these samples for manual labeling.

We are presented with two choices for labeling: (1) labeling the image and the caption, and (2) labeling only the image. We opted for (2) because we cannot guarantee caption veracity, especially considering the transformations they underwent during data collection of CC3M~\cite{sharma2018conceptual}.

We use LabelStudio\footnote{\url{https://labelstud.io/}}, an open-source data labeling platform, to label the images.
The question to be answered by the annotators is whether the image contains a child, with ``Child'' and ``NoChild'' as the two possible labels.
The authors agreed on five rules for the labeling task.

\textit{1. A child is defined as a person under 18 years of age.} Since we aim to detect depictions of minors, and in most jurisdictions, the majority age is 18, we stick to this definition of a child.

\textit{2. The annotator should label the image as ``Child'' if they believe it is more likely than not (i.e., more than 50\% chance) that one or more people in this image are children.} We thus consider only two possible labels during the labeling process, and annotators must select one of the two options.

\textit{3. The annotator must base their decision on the apparent age of the person.} Since in many jurisdictions apparent age is included in the definition of CSAM, the annotator should not try to find the ground truth (for example by using the search engines on the photo).
They should base their decision on the given image only. 
This also ensures that the same approach is applied to both real and fictional images.

\textit{4. Any depiction of a human, such as sculptures, drawings, and cartoon characters, is a candidate for the ``Child'' label.} Indeed, in some jurisdictions relevant laws apply to fictional depictions too~\cite{legal_fic_csam}
, in addition to photographs.

\textit{5. Partially visible bodies or images with poor quality should still be labeled if the annotator can identify a child.} Similarly to rules 3 and 4, legislation often does not make a difference with respect to quality of the images, or visibility of the face.

\begin{figure}[htb]
    \centering

    \begin{subfigure}{0.22\textwidth}
        \includegraphics[width=\linewidth,height=\linewidth]{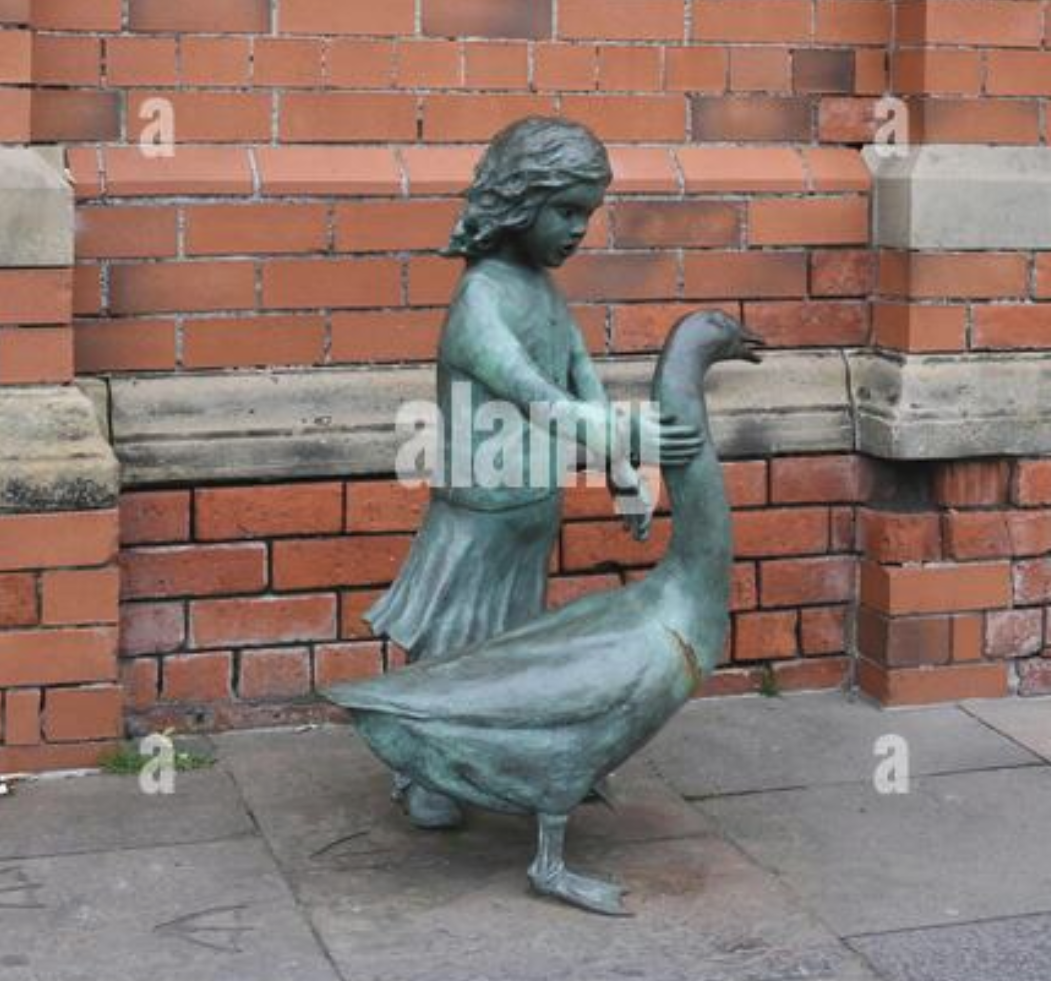}
        \caption{}
        \label{fig:image1}
    \end{subfigure}
    \hfill
    \begin{subfigure}{0.22\textwidth}
        \includegraphics[width=\linewidth,height=\linewidth]{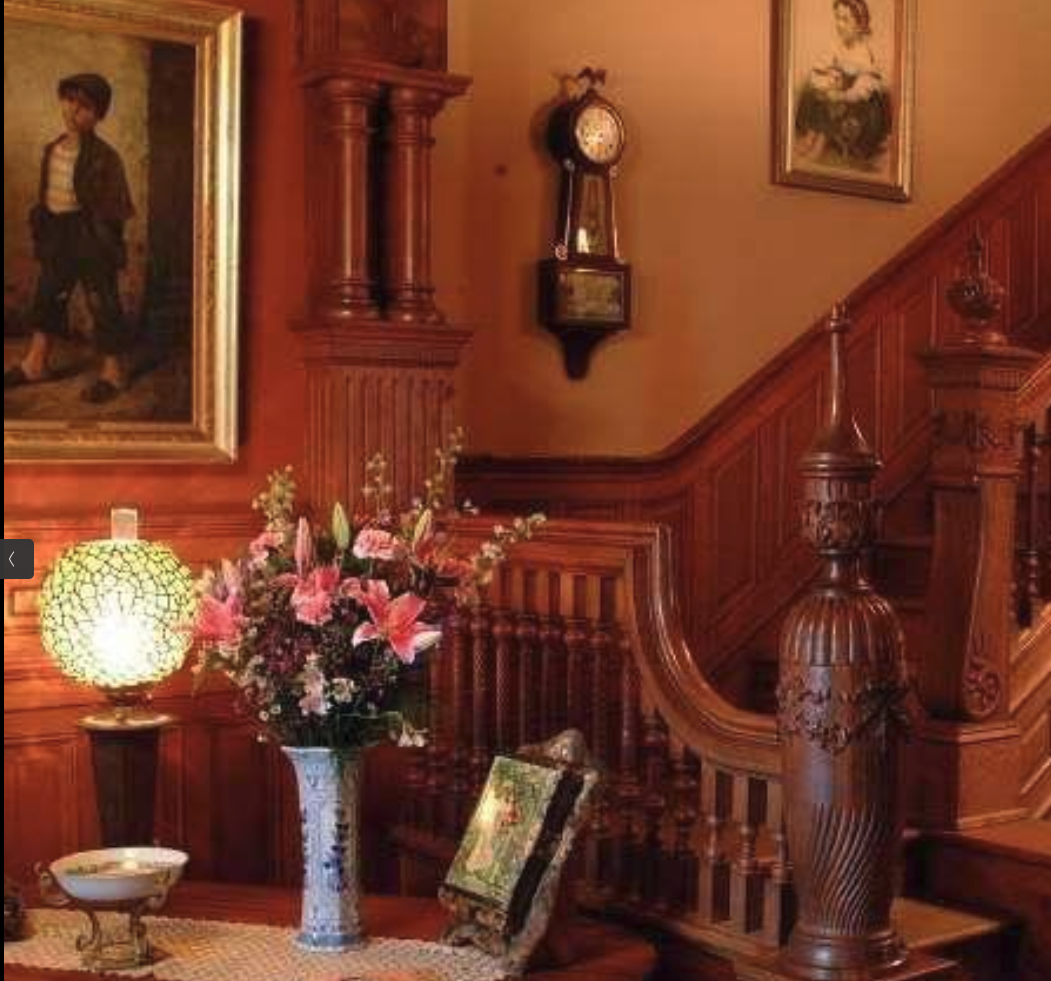}
        \caption{}
        \label{fig:image2}
    \end{subfigure}
    \hfill
    \begin{subfigure}{0.22\textwidth}
        \includegraphics[width=\linewidth,height=\linewidth]{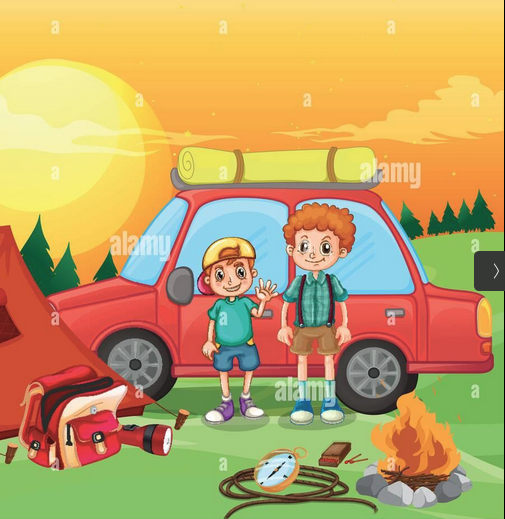}
        \caption{}
        \label{fig:image3}
    \end{subfigure}
    \hfill
    \begin{subfigure}{0.22\textwidth}
        \includegraphics[width=\linewidth,height=\linewidth]{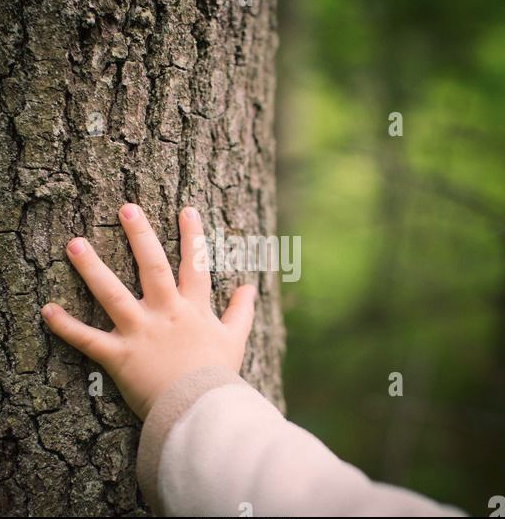}
        \caption{}
        \label{fig:image4}
    \end{subfigure}

    \begin{subfigure}{0.22\textwidth}
        \includegraphics[width=\linewidth,height=\linewidth]{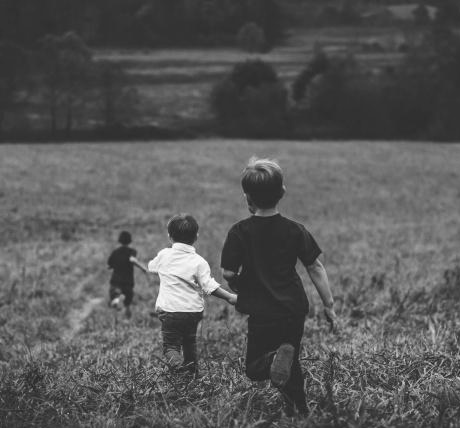}
        \caption{}
        \label{fig:image5}
    \end{subfigure}
    \hfill
    \begin{subfigure}{0.22\textwidth}
        \includegraphics[width=\linewidth,height=\linewidth]{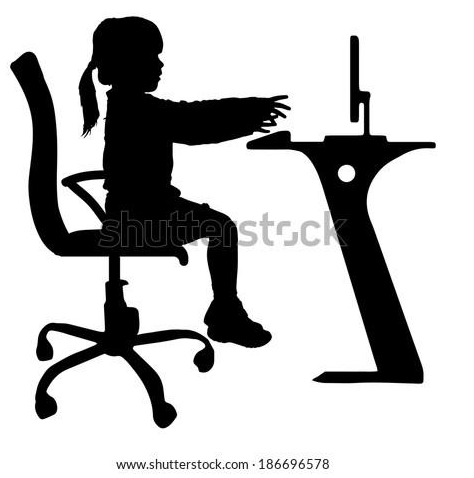}
        \caption{}
        \label{fig:image6}
    \end{subfigure}
    \hfill
    \begin{subfigure}{0.22\textwidth}
        \includegraphics[width=\linewidth,height=\linewidth]{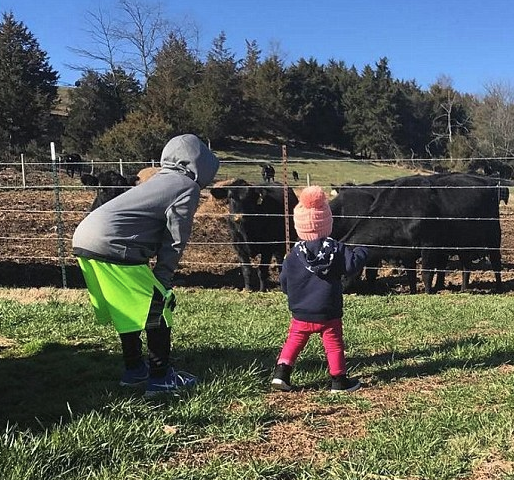}
        \caption{}
        \label{fig:image7}
    \end{subfigure}
    \hfill
    \begin{subfigure}{0.22\textwidth}
        \includegraphics[width=\linewidth,height=\linewidth]{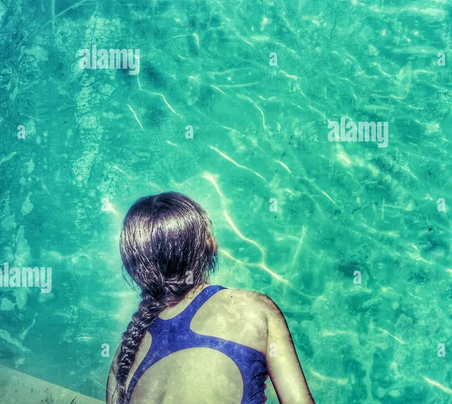}
        \caption{}
        \label{fig:image8}
    \end{subfigure}

    \begin{subfigure}{0.22\textwidth}
        \includegraphics[width=\linewidth,height=\linewidth]{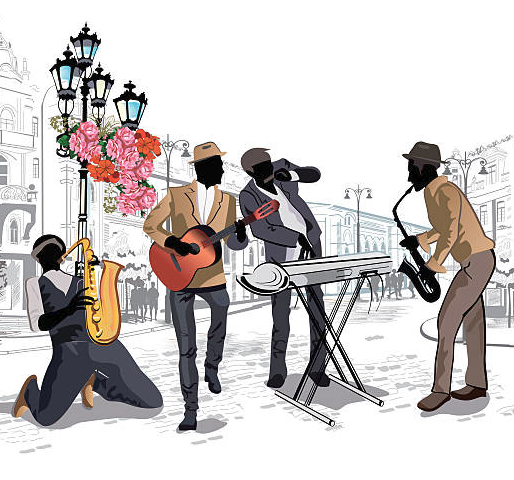}
        \caption{}
        \label{fig:image9}
    \end{subfigure}
    \hfill
    \begin{subfigure}{0.22\textwidth}
        \includegraphics[width=\linewidth,height=\linewidth]{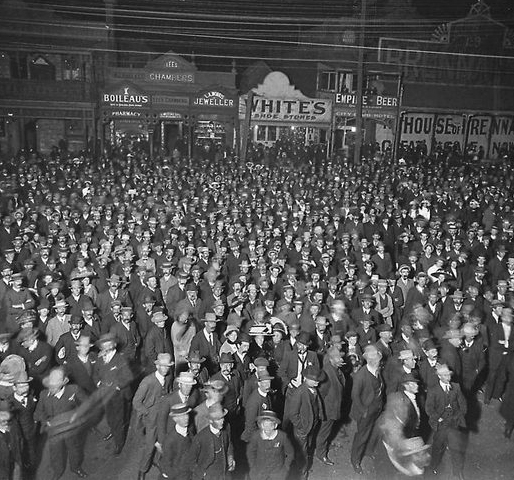}
        \caption{}
        \label{fig:image10}
    \end{subfigure}
    \hfill
    \begin{subfigure}{0.22\textwidth}
        \includegraphics[width=\linewidth,height=\linewidth]{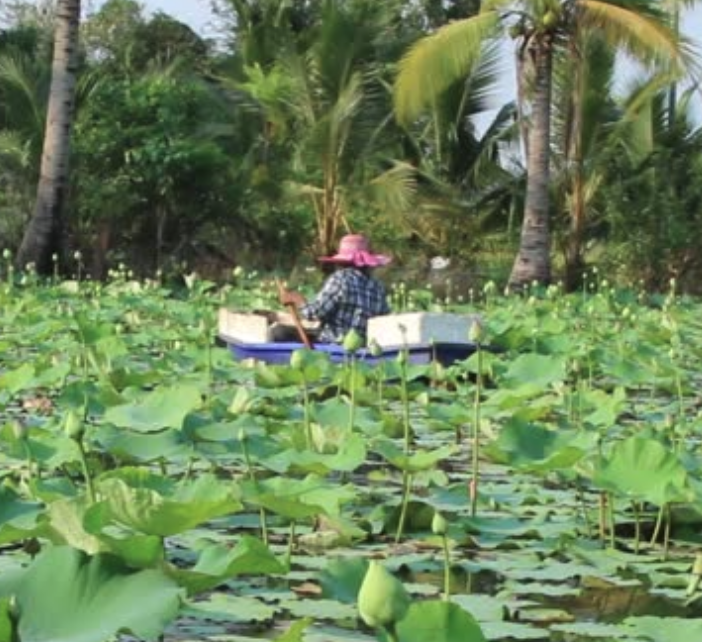}
        \caption{}
        \label{fig:image11}
    \end{subfigure}
    \hfill
    \begin{subfigure}{0.22\textwidth}
        \includegraphics[width=\linewidth,height=\linewidth]{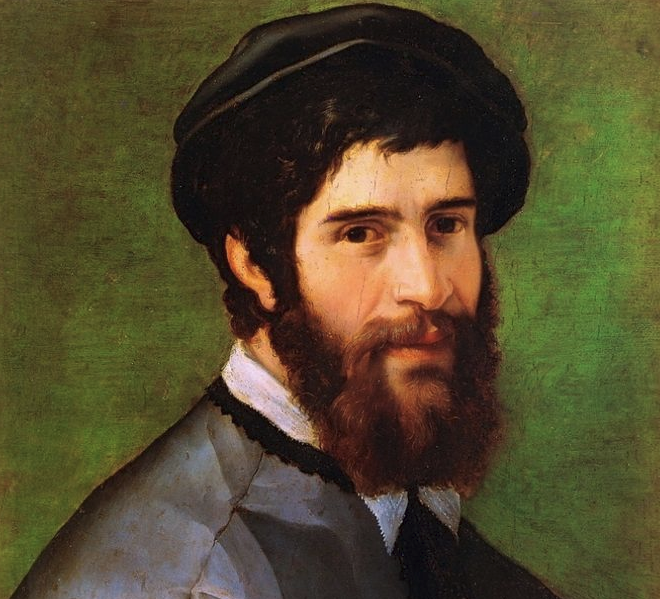}
        \caption{}
        \label{fig:image12}
    \end{subfigure}

    \caption{Examples of the images in our dataset. We deliberately do not show images with faces, for privacy reasons. Images \ref{fig:image1} - \ref{fig:image7} are labeled as "Child", Image \ref{fig:image8} is an example of a disagreement, and images \ref{fig:image9}-\ref{fig:image12} and labeled as "NoChild".}
    \label{fig:dataset-examples}
\end{figure}

Fig.~\ref{fig:dataset-examples} illustrates with examples how we applied these rules.

Two of the authors labeled all the images.
After labeling the first 1000 images separately, the two annotators discussed the disagreements. 

There were two types of disagreements.
First, disagreements due to mistakes by one of the annotators, e.g., when one of the annotators did not notice a child image on the background (Figure ~\ref{fig:image2}), these mistakes were fixed and we assigned ``Final\_Child'' or ``Final\_NoChild'' label.
Second, disagreements due to different opinions on the estimated age of a person/people. When a disagreement could not be resolved, the image was labeled as ``Disagreement''  (an example of ``Disagreement'' image is Figure ~\ref{fig:image8}). We exclude the ``Disagreement'' images from the evaluation in Section \ref{sec:filters}.

After this discussion, the authors separately labeled the rest of the images and resolved the new disagreements in the same way.

The process resulted in 1675 images labeled as containing a child, 8262 images labeled as not containing any child, and 63 ``Disagreement'' images.
Of the images that the annotators agreed upon, 16.9\% contain a child. This remarkable level of agreement (Cohen-Kappa score $\sim$0.98) indicates that the annotation task is well defined, and weakly depends on the annotator's personal opinion.

\textbf{Hashes.} As we do not control the URL domain where images are made available, it is possible for images to become unavailable at a later time. To avoid integrity issues, similarly to Carlini et al.~\cite{carlini2024poisoning}, we also release the hashes of the images so that users of the dataset can verify that the images they download are the same as the ones we labeled.
We use two types of hashes: perceptual PDQ hash (we use its Python implemention\footnote{\url{https://pypi.org/project/pdqhash}}) and SHA256. Since SHA256 is sensitive to any change within the image, and such changes may occur due to the use of a different downloading method, we assume that images with changed SHA256 hashes can still be used to enrich the dataset, as long as the distance between PDQ hashes remains small~\cite{jain2022adversarial}.

\section{Benchmarking Examples}
\label{sec:filters}
In this section, we evaluate how off-the-shelf solutions can perform on our benchmark, exploring both the image and caption modalities.

More specifically, we use Amazon Rekognition Image's face-based age estimator~\cite{amazonrekognition} as an \textit{image-based minor detector},
and DeepSeek's LLM~\cite{deepseek} as a \textit{caption-based minor detector}.

\subsection{Experimental Setup}

\textbf{Metrics.} Considering the unbalanced nature of our dataset, we report the following metrics: \textit{True Positive Rate (TPR)}, defined as the proportion of child images identified as a child image by the method; and \textit{False Positive Rate (FPR)}, defined as the proportion of images without a child identified as a child image by the method.

\textbf{Methods.} Our \textit{caption-based detector} uses DeepSeek-V3~\cite{liu2024deepseek}, a state-of-the-art large language model (LLM) developed by the DeepSeek company and accessed via the API.~\footnote{\url{https://api-docs.deepseek.com/}}
We label each caption individually using the prompt given in Appendix \ref{app:prompt} using default parameters for LLM generation.

Our \textit{image-based detector} is based on Amazon Rekognition Image, a service providing API access to a DetectFaces functionality.\footnote{\url{https://docs.aws.amazon.com/rekognition/latest/dg/faces-detect-images.html}}
DetectFaces takes as input an image and returns inferred face attributes for the 100 largest faces detected in the image, including age range.

For every image in the dataset, we use DetectFaces to retrieve the age ranges of faces $F_1, \ldots, F_n$ detected in the image, denoted as $[l_i, h_i]$ for the $i$-th face $F_i$.
The age range signifies that, according to DetectFaces, the individual with face $F_i$ is between $l_i$ and $h_i$ years old.

If no face is detected ($n=0$), we return False, meaning that no child is detected.
If at least one face is detected ($n>0$), we propose two different classification rules to determine if the image contains a child. Let $[(l_1, h_1), \ldots, (l_n, h_n)]$ denote the list of age ranges of $n$ faces detected in the image and let $\tau$ be a threshold (a typical value is $\tau=18$ corresponding to the age of majority).
Our two rules are:
\begin{enumerate}
    \item Min-range rule: return \textit{Child} if and only if $\min\limits_{i=1,\ldots,n}l_i < \tau$.
    \item Mid-range rule: return \textit{Child} if and only if $\min\limits_{i=1, \ldots, n}\frac{l_i+h_i}{2}<\tau$.
\end{enumerate}

In order to build a detector with better TPR, the two methods can also be combined to form an \textit{image-caption-based detector}. More specifically, we combine their outputs are combined via logical OR, i.e. Child | Non-Child -> Child. 

\textbf{Compute requirements.} 
Evaluating the methods does not require specific compute, as we query external APIs. 
The financial cost of our experiments was 0.12\$ paid to DeepSeek API and 12\$ paid to Amazon Rekognition Image. The Deepseek experiments can take 1-13 hours, depending on the number of processes used to parallelize the requests and the time of the day (requests are processed faster during off-peak hours), and Amazon Rekognition Image experiments take 1-2 hours. Any machine with 8 or more CPU cores should be sufficient to run the experiments.

\subsection{Experimental Results}
We evaluate each method on our dataset.
Table~\ref{tab:results} suggests that minor detection is a challenging task on our dataset, both in the caption and image modalities.

\begin{figure}[htbp]
    \centering

    \begin{subfigure}{0.33\textwidth}
        \centering\includegraphics[width=\linewidth,height=0.8\linewidth]{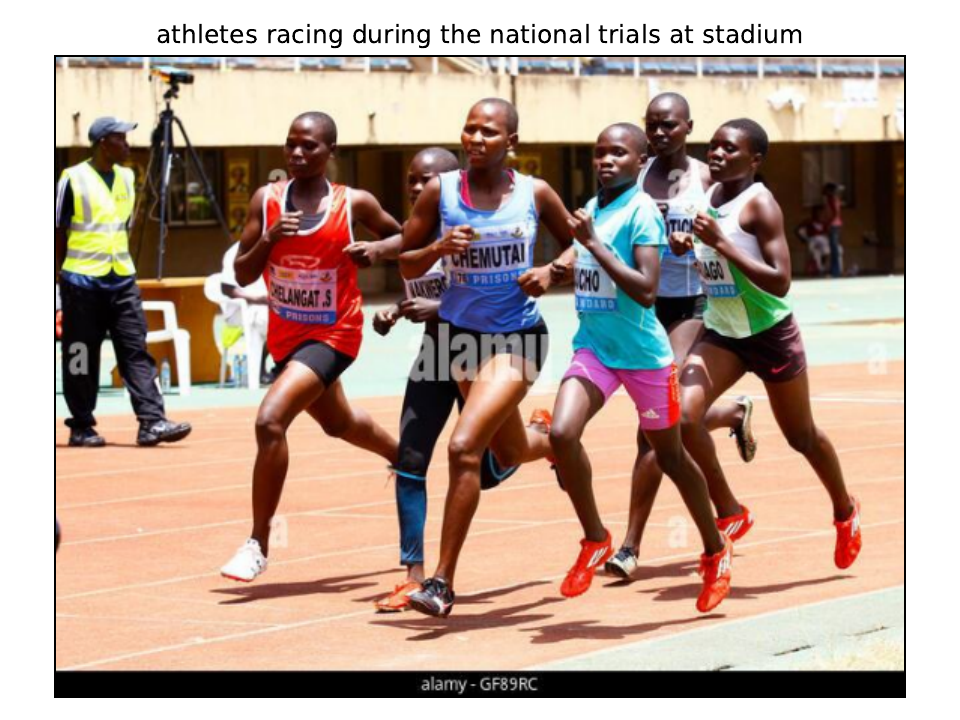}
        \label{fig:fn-deepseek-1}
    \end{subfigure}
    \hfill
    \begin{subfigure}{0.33\textwidth}
        \centering\includegraphics[width=\linewidth,height=0.8\linewidth]{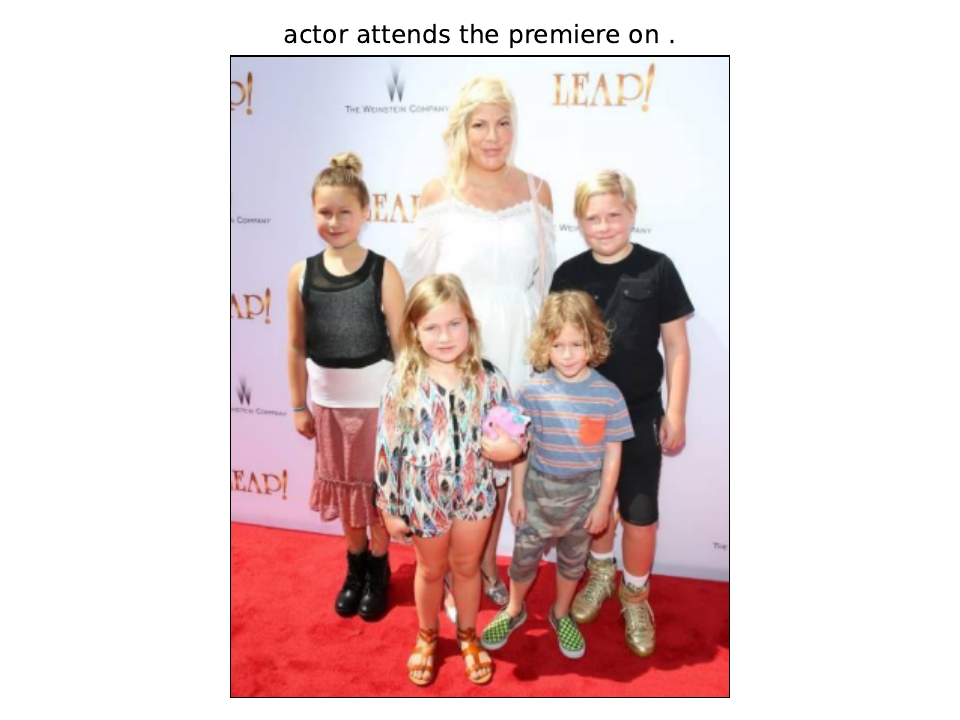}
        \label{fig:fn-deepseek-3}
    \end{subfigure}
    \begin{subfigure}{0.32\textwidth}
        \centering\includegraphics[width=\linewidth,height=0.9\linewidth]{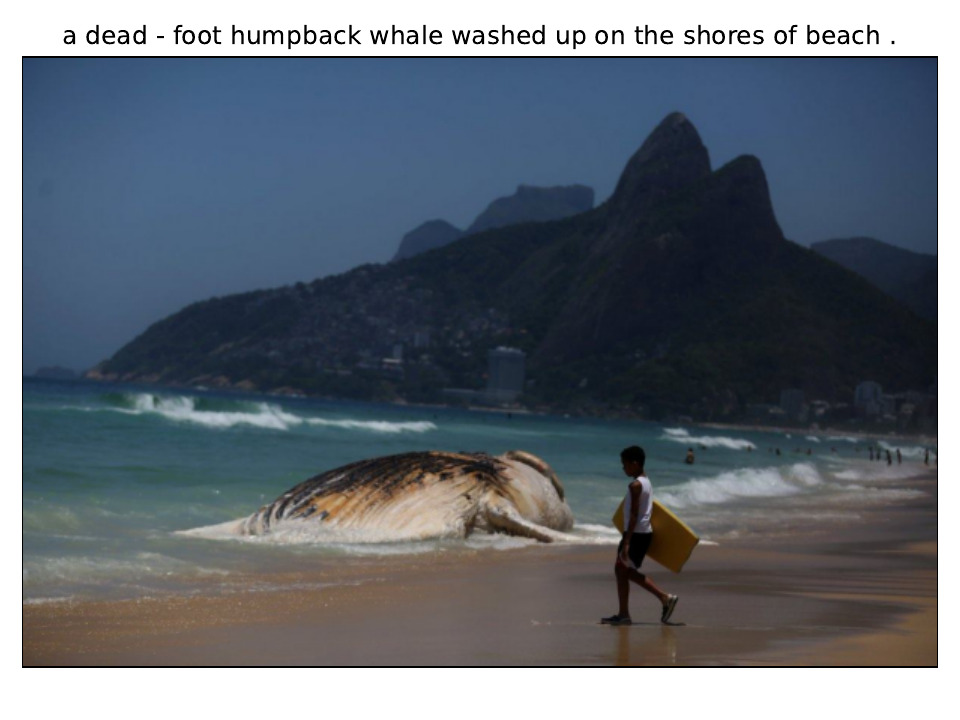}
        \label{fig:fn-deepseek-4}
    \end{subfigure}

 \begin{subfigure}{0.32\textwidth}
        \centering\includegraphics[width=0.85\linewidth,height=\linewidth]{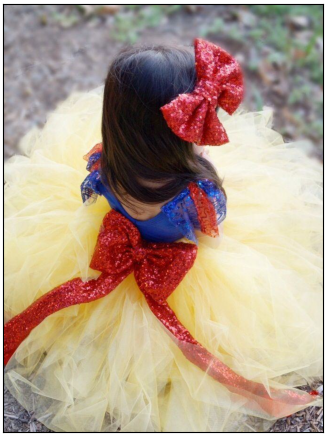}
        \label{fig:fn-amazon-1}
    \end{subfigure}
    \hfill
    \begin{subfigure}{0.32\textwidth}
        \centering\includegraphics[width=0.85\linewidth,height=0.85\linewidth]{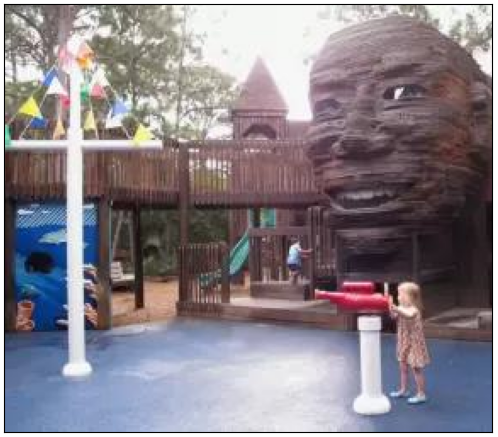}
        \label{fig:fn-amazon-2}
    \end{subfigure}
     \begin{subfigure}{0.32\textwidth}
        \centering\includegraphics[width=0.85\linewidth,height=0.85\linewidth]{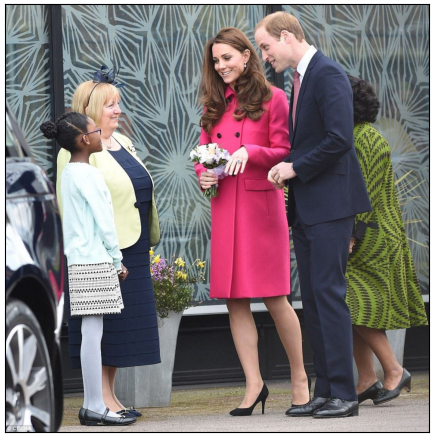}
        \label{fig:fn-amazon-3}
    \end{subfigure}
   
    \caption{Examples of false negatives of caption-based minor detection using DeepSeek API (top row) and Amazon Image Rekognition (bottom row). 
    DeepSeek false negatives are images with children, where the captions do not suggest their presence, and are therefore not flagged by DeepSeek. For Amazon, false negatives are images with children, where the face is not visible or only partially visible, resulting in no face being detected or age estimation being inaccurate.}
    \label{fig:false-negatives}
\end{figure}

\textbf{Caption-based minor detection} using DeepSeek-V3 achieves a relatively low TPR of only 45.9\%.
We attribute this low TPR to two reasons.
The first reason is that the method may not correctly identify all images that refer to children even though the captions explicitly refer to children.
A manual inspection of the false negatives reveals several captions for which this occurs, for instance ``students share their happiness with their teacher'', ``boys and a girl roll a giant snowball date'', ``group of high school students standing by locker''.
It also reveals captions that are more ambiguous yet still strongly suggest the presence of a child, e.g., ``happy family on the beach''.
The second reason is that many captions do not make any reference to children nor to situations where a child is likely to present, even if a child is present in the image, e.g., ``athletes racing during the national trials at stadium'', ``actor attends the premier on'', and ``a dead-foot humpback whale washed up on the shores of beach'', shown in the top row of Fig.~\ref{fig:false-negatives}.
\begin{table*}[htbp]
  \centering
  \begin{tabular}{lccc} 
  \toprule
    \textbf{Modality} & \textbf{Method} & \textbf{TPR} & \textbf{FPR}  \\
  \midrule
    Caption & DeepSeek-V3~\cite{liu2024deepseek} & 45.9\% & 3.3\% \\
  \midrule
    \multirow{2}{*}{Image} & Amazon (Min-range rule, $\tau=18$) & 64.1\%  & 5.1\%  \\
    & Amazon (Mid-range rule, $\tau=18$) & 61.0\%  & 2.5\%  \\
    \midrule
    \multirow{2}{*}{Image and caption} & DeepSeek-V3 + Amazon (Min-range rule, $\tau=18$) & 75.3\%  & 8.2\% \\
     & DeepSeek-V3 + Amazon (Min-range rule, $\tau=18$) & 72.5\%  & 5.6\% \\
 \bottomrule \\
  \end{tabular}
  \caption{Result of off-the-shelf minor detection methods on our dataset. For space reasons, Amazon Rekognition Image is referred to more briefly as Amazon.}
  \label{tab:results}
\end{table*}
This suggests that even in a highly curated dataset like CC3M, captions alone are not informative enough to correctly identify all images with minors.
On the other hand, caption-based classification using  DeepSeek-V3 is very good at identifying captions that do not refer to children, achieving a very low FPR of 3.3\%.
We attribute this to the LLM answering ``no'' whenever the caption does not explicitly mention children, as is the case in most samples without any children.
Thanks to its low FPR, caption-based classification may therefore still present advantages if it is able to identify images of minors on which image-based classification fails, as we show later.
Fig.~\ref{fig:false-positives-deepseek} in the Appendix \ref{app:fp} shows some examples of false positives.

\textbf{Image-based minor detection} using Amazon Rekognition Image achieves a much higher TPR, of 64.1\% using the min-range rule and of 61.0\% using the mid-range rule, both with a threshold $\tau=18$.
The success of the former can be attributed to using a more conservative rule than the latter for age classification, at the cost of a higher FPR (5.1\% instead to 2.5\%).
The best TPR achieved is, however, far from perfect, leaving many images of children undetected, and highlighting the limitations of face-based age estimation methods for the minor detection task.
We will, from now on, focus on Amazon Rekognition Image using the min-range rule as the default, given its superior performance.

\begin{figure}[htp]
    \centering
    \includegraphics[scale=0.6]{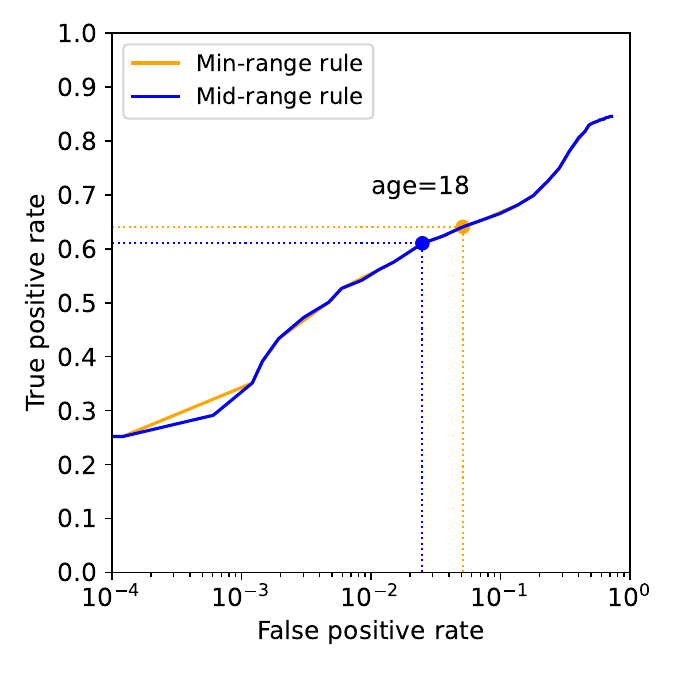}
    \caption{Receiver Operating Characteristics (ROC) curve of minor detection methods using Amazon Rekognition Image on our dataset. We plot the TPR vs. FPR for different values of the threshold $\tau$.}
    \label{fig:amazon-roc}
\end{figure}

Fig.~\ref{fig:amazon-roc} shows that the TPR of Amazon Rekognition Image can be increased to 84.5\% by increasing the age threshold $\tau$. 
It cannot be increased further, due to the method not detecting faces in 15.5\% of the images.
However, this increase comes at a steep increase in the FPR to 72.1\%. 
Even a small increase in the TPR, from 64.1\% to 74.9\%, would lead to a 5.5$\times$ increase in the FPR.

\begin{figure}[htbp]

    \begin{subfigure}{0.3\textwidth}
        \centering\includegraphics[width=\linewidth,height=0.8\linewidth]{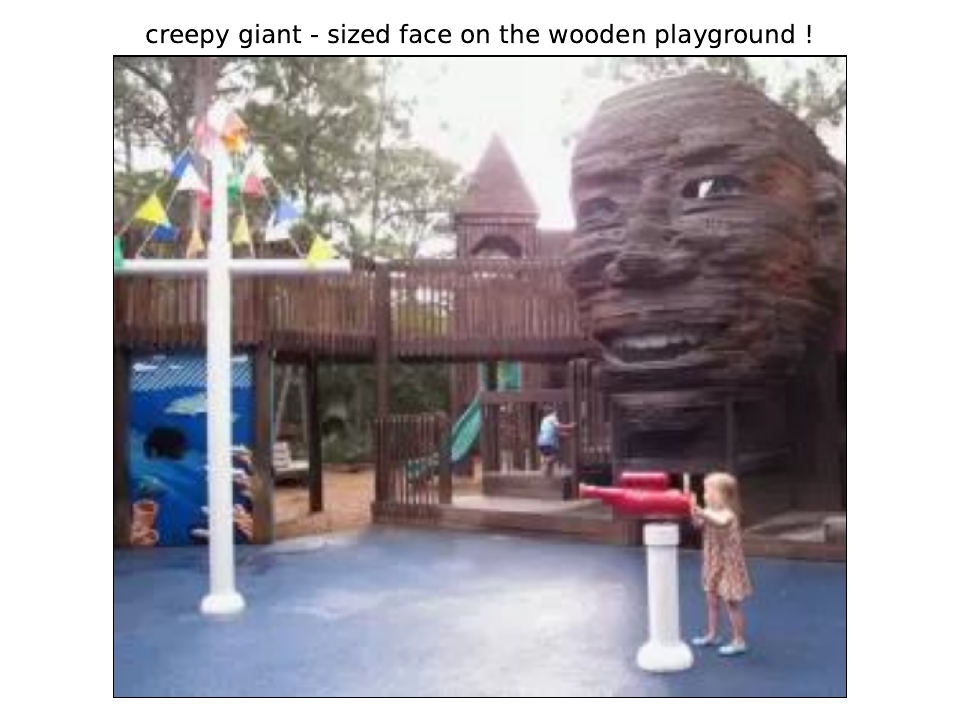}
        \label{fig:fp-amazon-2}
    \end{subfigure}
    \hfill
    \begin{subfigure}{0.3\textwidth}
        \centering\includegraphics[width=\linewidth,height=0.8\linewidth]{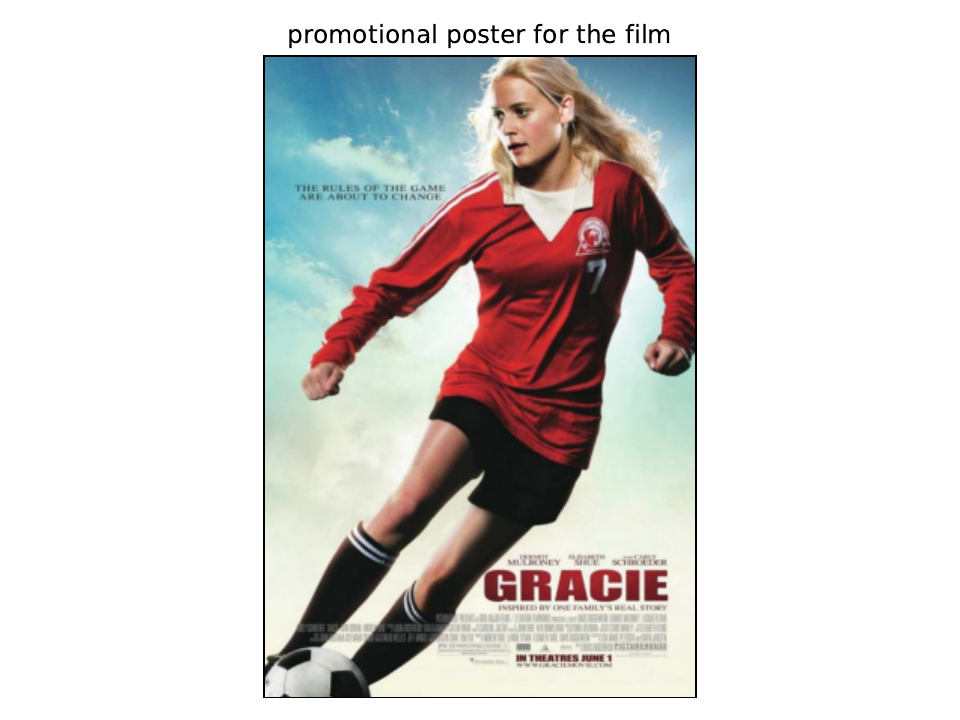}
        \label{fig:fn-combined-3}
    \end{subfigure}
    \hfill
    \begin{subfigure}{0.3\textwidth}
        \centering\includegraphics[width=\linewidth,height=0.85\linewidth]{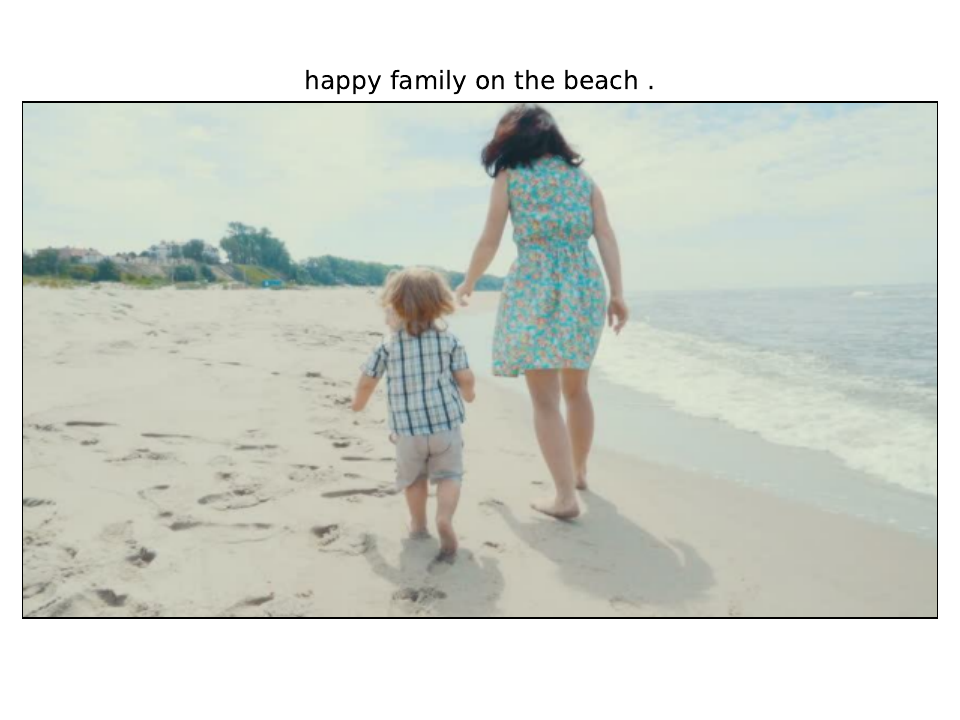}
        \label{fig:fn-combined-4}
    \end{subfigure}

    \caption{Examples of false negatives of the combined method. 
    These are images with children whose captions either suggest the presence of a child but are incorrectly classified by DeepSeek or do not suggest the presence of a child; and on which Amazon Image Rekognition fails to detect a child, either because the face of the child is not visible and therefore not detected or because age estimation is inaccurate on detected faces.}
    \label{fig:false-negatives-combined}
\end{figure}

Fig.~\ref{fig:false-negatives} (bottom row) shows examples of false negatives produced by Amazon Rekognition Image.
Some of the false negatives are images where the face of the child is not visible, or the face of the child is not detected, either because it is too small relative to the size of the image, or because the image is a fictional depiction of a child. Fig.~\ref{fig:false-positives-amazon} in Appendix \ref{app:fp} shows examples of false positives of Amazon Rekognition Image. We find many images where age estimation is inaccurate. %

\textbf{Image-caption-based minor detection.} An analysis of mistakes made by caption- and image-based minor detection methods reveals how the two approaches can complement each other.

We find that 11.2\% of child images are correctly detected by the DeepSeek API but not by Amazon Rekognition Image, and 29.4\% of child images are correctly detected by Amazon Rekognition Image but not by the DeepSeek API.  
The combined classifier thus correctly identifies 75.3\% of child images, with an FPR of 8.2\% 
In general, the caption-based classifier can flag many images where the child is mentioned in the caption but the face is not visible in the image, while the image-based classifier can flag images where the face is visible but the child is not mentioned in the caption.
None can flag images where the face of the child is not visible in the image and the child is not mentioned in the caption, or the caption is not suggestive of a child's presence. Fig.~\ref{fig:false-negatives-combined} shows such false negative examples, including one example where the caption suggests the presence of children, referring to ``playground'', and one example where the face of the minor is visible but age estimation failed.

\section{Conclusion, Limitations, and Future Work}
\label{sec:conc}
Most digital platforms restrict the sharing of content that relates to minors, and minor detection methods based on machine learning are used to detect such content. 
In this paper, we present Image-Caption Children in the Wild Dataset (ICCWD), the first image-caption dataset for benchmarking minor detection methods in a multi-modal environment using caption in addition to image information.
ICCWD is richer than previous image datasets for minor detection, containing images of children in a variety of contexts, including fictional depictions and partially visible bodies where the face is not visible.
We demonstrate the utility of our dataset by benchmarking three minor detection methods and show that minor detection is a challenging task. We hope our dataset will aid in the design of better minor detection methods.

\textbf{Limitations and Future Work.} One limitation of our work is the size of the dataset. While we find the resulting number of images enough for the benchmarking purposes, a larger dataset could enable more ambitious tasks such as training a minor detection model. Due to the manual labeling procedure, we do not find this extension feasible.

Another limitation is that we have used only two annotators to create our dataset. Such an approach has benefits -- it allowed for us to discuss and resolve nearly all disagreements and achieve a high degree of consistency in our ratings. Yet, due to the intrinsic ambiguity of the annotation task we could not resolve our disagreements on all images. Disagreement-labeled images, while currently not used, could be relabeled in the future.

\section{Ethical considerations}
\label{sec:ethics}

\textbf{Dataset potential harmful uses.} 
Our dataset contains 1675 children images. Someone could use these images for malicious purposes, for instance to fine-tune a T2I model on the likeness of children. 
We believe that, given the small volume of images of children relative to existing datasets, the increase in risk resulting from publishing this dataset is marginal compared to other publicly available child datasets~\cite{gangwar2021attm,phsmoura2020,ylfw2024,falkenberg2024syn,styleganchilddataset} that 
contain more images of children than ours, making them more suitable for such purposes. 

Yet, we acknowledge that providing a benchmark to improve minor detection methods in image-caption datasets may allow to build larger datasets of images depicting minors that could be used for fine-tuning (or could be in themselves of interest for illicit purposes). 
While this is a risk, we believe that the societal benefits associated with  enabling the improvement of minor detection methods, including methods for filtering depictions of children before training text-to-image (T2I) models at large scale, e.g., to prevent these models from reproducing the likeness of children for privacy reasons~\cite{hrw2024australia} and to potentially prevent AI-CSAM generation by these models~\cite{thorn2024}, is greater than the potential negative consequences -- given that criminals might already have such  datasets of minor pictures, and can use other methods, including manual labeling, to obtain enough images for fine-tuning.

\textbf{Privacy of people in the dataset.} Our dataset contains images of people, including minors, that are publicly available at the URLs released in the CC3M dataset~\cite{sharma2018conceptual}.
Similarly to other image-caption datasets~\cite{sharma2018conceptual,schuhmann2022laion}, we only release the URLs, alongside a script for downloading images from the URLs.
This allows people included in the dataset to take down their images from the URLs, making them unavailable for future downloads.
A downside of only releasing URLs is that future users of the dataset may not be able to download all the images originally labeled.
However, we believe that the privacy benefits of only releasing URLs outweigh the downside of future works evaluating their minor detection methods on slightly different versions of the dataset.

This paper includes some images with real people, for the purposes of illustrating our methodology (Fig.~\ref{fig:dataset-examples}) and showing examples of errors made by the different minor detection methods (Fig.~\ref{fig:false-negatives}-~\ref{fig:false-positives-amazon}).
We have exclusively selected  images of public events (e.g., concert, film premiere event) and public people (e.g., actors, royalty), and images where the face of the minor is not visible.

\textbf{License of source dataset.} Our dataset is sourced from Google's CC3M~\cite{sharma2018conceptual}. Our use of CC3M is compliant with its license,\footnote{\url{https://github.com/google-research-datasets/conceptual-captions/blob/master/LICENSE}} (see Appendix~\ref{app:license} for details).

\section{Acknowledgements}
This research was partially supported by armasuisse Science and Technology, and as part of the Swiss AI Initiative by a grant from the Swiss National Supercomputing Centre (CSCS) under project ID a07 on Alps.

\bibliographystyle{plain}
\bibliography{bibliography}


\appendix

\section{License of source dataset}\label{app:license}

Our dataset is sourced from Google's CC3M dataset~\cite{sharma2018conceptual}, which allows for downstream uses of the dataset such as ours.
The license states: ``The dataset may be freely used for any purpose, although acknowledgement of
Google LLC ("Google") as the data source would be appreciated. The dataset is
provided "AS IS" without any warranty, express or implied. Google disclaims all
liability for any damages, direct or indirect, resulting from the use of the
dataset.''

\section{Technical Appendices and Supplementary Material}
\subsection{DeepSeek Prompt}
We provide below the prompt we used to label captions using DeepSeek-V3~\cite{liu2024deepseek}.
\label{app:prompt}
\begin{tcolorbox}[width=\textwidth,title={\textbf{Prompt and code used to label captions using DeepSeek API}},outer arc=0mm]    
\texttt{prompt="You are a helpful assistant. Does this caption refer to a }

\;\; \texttt{child? You must answer with yes or no."}

\texttt{response = client.chat.completions.create(\{}

\;\; \texttt{model="deepseek-chat",}

\;\;            \texttt{messages=[}
            
\;\;\;\;                \texttt{\{"role": "system", 
                "content": prompt\},}
                
\;\;\;\;                \texttt{\{"role": "user", "content": "Caption: " + caption\}},
                
\;\;                \texttt{],}
                
\;\;             \texttt{max\_tokens=4,}
            
\;\;             \texttt{stream=False]}\})
\end{tcolorbox} 

\subsection{False Positive Error Samples}
\label{app:fp}
\begin{figure}[htbp]
    \centering

    \begin{subfigure}{0.45\textwidth}
        \centering\includegraphics[width=\linewidth,height=0.8\linewidth]{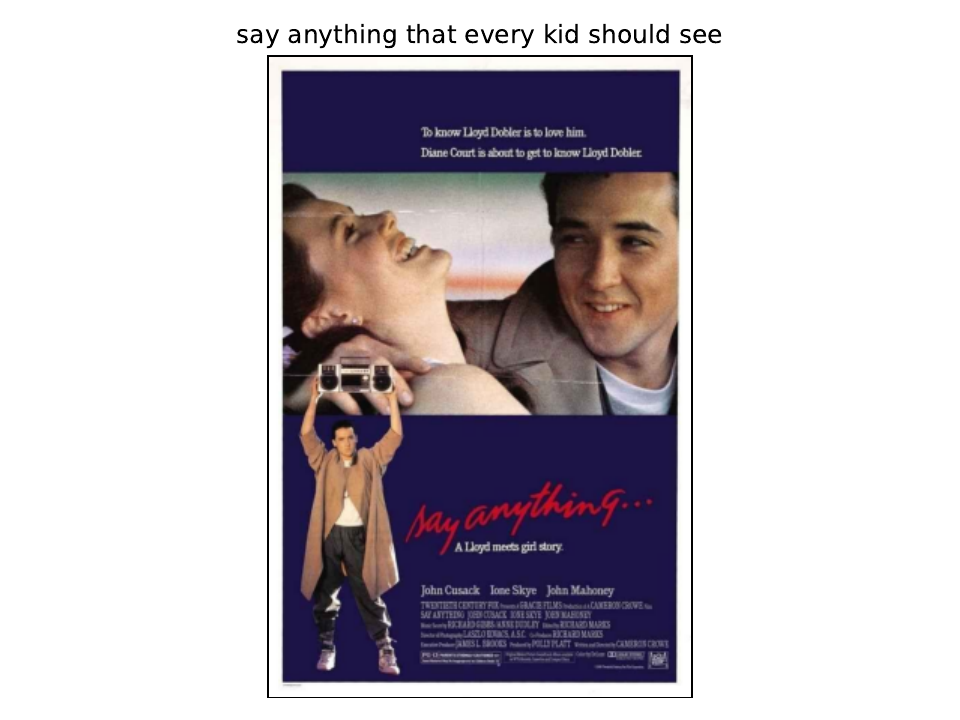}
        \label{fig:fp-deepseek-1}
    \end{subfigure}
    \hfill
    \begin{subfigure}{0.45\textwidth}
        \centering\includegraphics[width=\linewidth,height=0.8\linewidth]{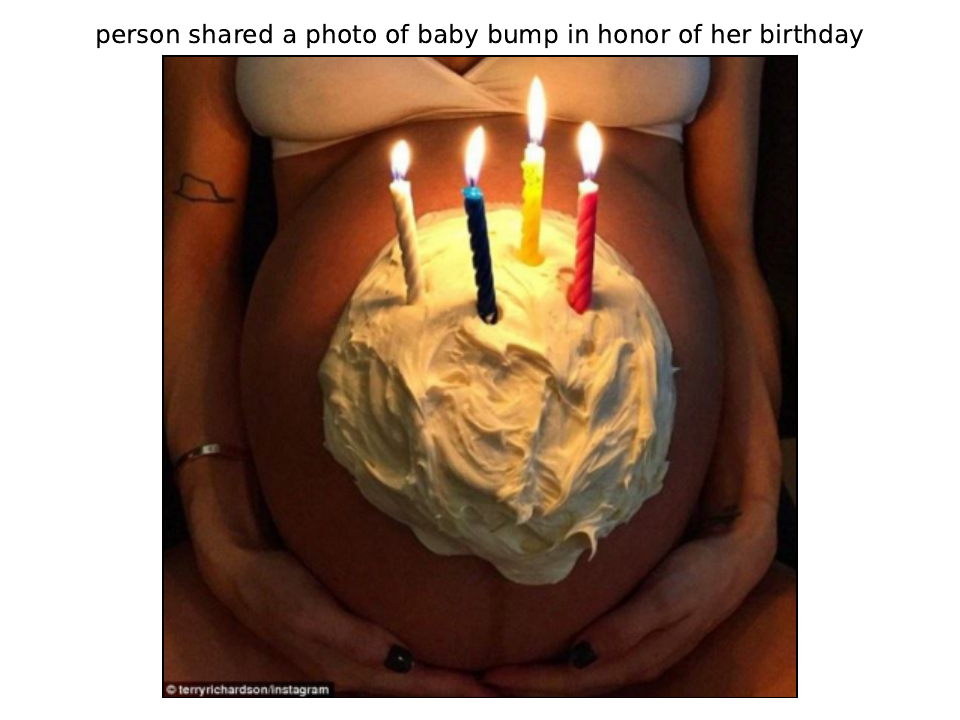}
        \label{fig:fp-deepseek-3}
    \end{subfigure}
     \begin{subfigure}{0.6\textwidth}
        \centering\includegraphics[width=\linewidth,height=0.8\linewidth]{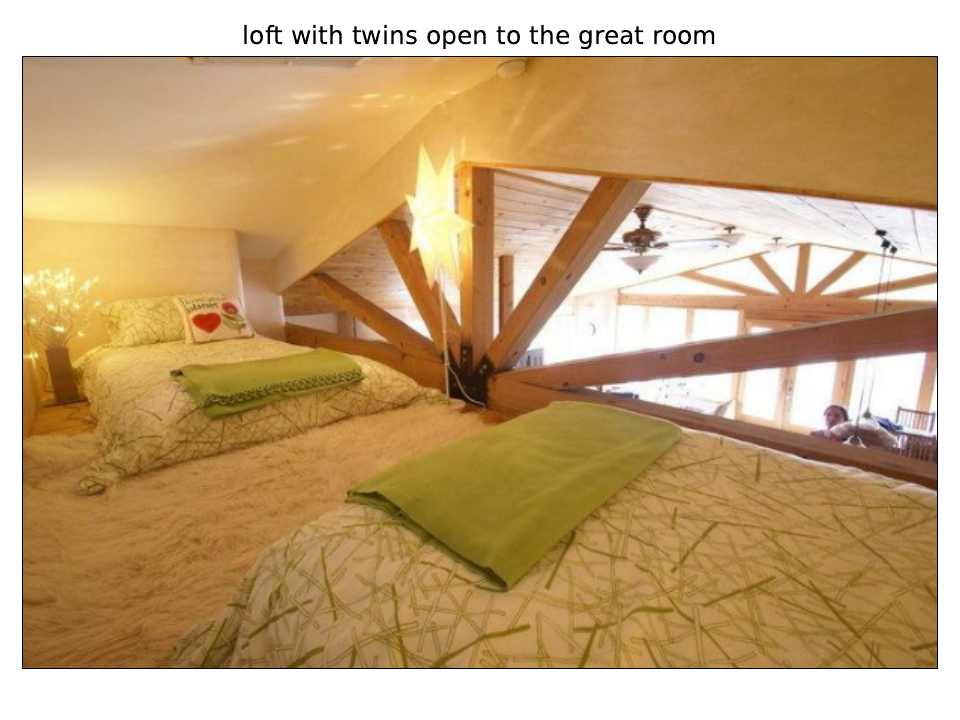}
        \label{fig:fp-deepseek-3}
    \end{subfigure}
   
    \caption{Examples of false positives of caption-based minor detection using DeepSeek API. 
    These are images without children that are erroneously flagged by DeepSeek, probably due to their captions containing child-related keywords.}
    \label{fig:false-positives-deepseek}
\end{figure}

\begin{figure}[htbp]
    \centering

    \begin{subfigure}{0.25\textwidth}
        \centering\includegraphics[width=0.7\linewidth,height=\linewidth]{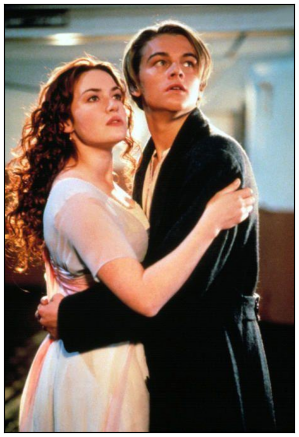}
        \label{fig:fp-amazon-1}
    \end{subfigure}
    \hfill
    \begin{subfigure}{0.25\textwidth}
        \centering\includegraphics[width=0.95\linewidth,height=0.95\linewidth]{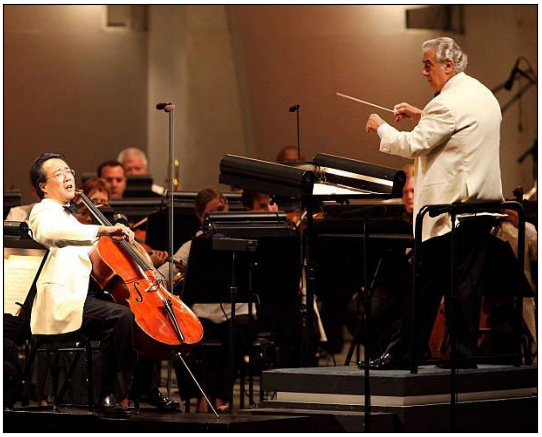}
        \label{fig:fp-amazon-2}
    \end{subfigure}
     \begin{subfigure}{0.45\textwidth}
        \centering\includegraphics[width=0.9\linewidth,height=0.6\linewidth]{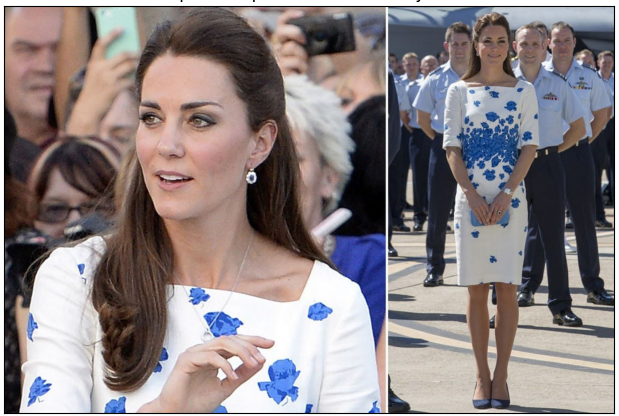}
        \label{fig:fp-amazon-3}
    \end{subfigure}
   
    \caption{Examples of false positives of image-based minor detection using Amazon Image Rekognition. 
    These are images without children on which age estimation is inaccurate.}
    \label{fig:false-positives-amazon}
\end{figure}

Fig.~\ref{fig:false-positives-deepseek}-~\ref{fig:false-positives-amazon} contain examples of false positives of our caption-based and image-based detection methods.

\end{document}